\documentclass[preprint,12pt]{elsarticle}

\usepackage{amssymb}

\usepackage{indentfirst}
\usepackage[usenames]{color}
\usepackage{algorithmic}
\usepackage{algorithm}
\usepackage{amsmath}
\usepackage{amsthm}
\usepackage{rotating}
\usepackage{graphics}
\usepackage{multirow}

\usepackage{longtable,lscape}
\usepackage[labelfont=bf]{caption} 
\usepackage{colortbl} 

\journal{Elsevier}

\begin{document}

\begin{frontmatter}



\title{A Memetic Algorithm for the Linear Ordering Problem with Cumulative Costs}


\author[HUST]{Tao Ye}
\ead{yeetao@gmail.com}
\author[HUST]{Kan Zhou}
\author[HUST]{Zhipeng L\"u\corref{cor}}
\ead{zhipeng.lui@gmail.com}
\cortext[cor]{Corresponding author.}
\author[Angers]{Jin-Kao Hao}
\ead{hao@info.univ-angers.fr}
\address[HUST]{School of Computer Science and Technology, Huazhong University of Science and Technology, 430074 Wuhan, P.R.China}
\address[Angers]{LERIA, Faculty of Sciences, University of Angers, 2, Boulevard Lavoisier, 49045 Angers, France}

\renewcommand{\baselinestretch}{1.5}\large\normalsize

\begin{abstract}
This paper introduces an effective memetic algorithm for the linear ordering problem with cumulative costs. The proposed algorithm combines an order-based recombination operator with \textcolor{black}{an improved forward-backward} local search procedure and employs \textcolor{black}{a solution quality based} replacement criterion for pool updating. Extensive experiments on 118 well-known benchmark instances show that the proposed algorithm achieves competitive results by identifying 46 new upper bounds. Furthermore, some critical ingredients of our algorithm are analyzed to understand the source of its performance.


\end{abstract}




\end{frontmatter}

\renewcommand{\baselinestretch}{1.5}\large\normalsize

\section{Introduction}
\label{introduction}
Given a complete directed graph $G=(V,E)$ with nonnegative vertex weight $d_i$ and nonnegative arcs cost $C_{ij}$, \textcolor{black}{where $V$ is the set of vertices ($n=|V|$)}, the Linear Ordering Problem with Cumulative Costs (LOPCC) aims to find a permutation $\pi=(\pi_1,\pi_2,\dots,\pi_n)$ of the $n$ vertices of $G$ such that the following function is minimized:
\theoremstyle{definition} \newtheorem{definition}{Definition}
\begin{equation} \label{eq_objective_function}
f(\pi)= \sum_{i=1}^n \alpha_{\pi_i}
\end{equation}
where
\begin{equation}
\alpha_{\pi_i} = d_{\pi_i}+\sum_{j=i+1}^nC_{\pi_i\pi_j}\alpha_{\pi_j}  \,\ \ \  \textrm{for} \ \ i=n, n-1, \dots, 1
\end{equation}

LOPCC was originally introduced in \cite{Bertacco} to formulate a practical problem appeared in wireless communication systems. Since then, a number of solution approaches have been proposed to solve this problem. Benvenuto \textsl{et al.} proposed a heuristic approach based on greedy construction \cite{Benvenuto}, which randomly selects a vertex from a reduced candidate list of the best available vertices and inserts the vertex into the \textcolor{black}{permutation} under construction. This algorithm is very fast, but its solution quality is generally unsatisfactory. Righini proposed an exact algorithm using branch-and-bound and a truncated branch-and-bound heuristic algorithm (TB\&B) \cite{Branch_Bound}. \textcolor{black}{The truncation technique can reduce the running time by a factor of $10$.} Despite of its promising results, the high computational time cost of TB\&B still prevented this algorithm from solving problems of size larger than $n=35$ in a reasonable time.

On the other hand, a number of metaheuristic algorithms were proposed to solve the LOPCC problem and proved to be effective to find high-quality solutions in a reasonable time. Some representative metaheuristic approaches include Iterated Local Search \cite{ILS1,ILS2}, Tabu Search \cite{TS},  Iterated Greedy-Strategic Oscillation and Path-Relinking \cite{EvPR}. Nevertheless, large instances (e.g., graphs with $n \geq 100$) still represent a real challenge for all existing LOPCC approaches.

This paper presents for the first time a memetic algorithm for solving the LOPCC problem. The proposed algorithm integrates a new local search procedure within the  framework of evolutionary computing. In particular, the proposed algorithm employs an order-based recombination operator and \textcolor{black}{a solution quality based} replacement strategy for population updating.

The performance of the proposed algorithm is assessed on a total of 118 LOPCC benchmark instances from the literature. Computational experiments show that the memetic algorithm matches the previous best solutions for 61 instances and improves the previous best solutions in 46 cases. To provide insights about the performance of the proposed algorithm, we analyze the influence of some critical ingredients of the algorithm.

The rest of this paper is organized as follows. Section 2 describes the key components of the memetic algorithm: The local search procedure, the recombination operator and the pool updating strategy. Section 3 gives the computational results and the comparison between the memetic algorithm and some state-of-the-art algorithms in the literature. Section 4 investigates several essential components of the proposed algorithm and conclusion is given in Section 5.

\section{Memetic algorithm}
\label{Sec_Algorithms}
\subsection{Main Scheme}
\label{main_scheme}

Memetic algorithms are known to be a powerful framework to solve hard combinatorial optimization problems (see e.g. \cite{Moscato1999,Hao2012}). By combining the evolutionary framework with the local search procedure, memetic algorithms are expected to offer a balance between an intensified examination of a given search region and an exploratory discovery of new promising areas.

Following the general design principle of memetic algorithms \cite{Hao2012}, our memetic approach alternates between a recombination phase to generate new solutions and a local optimization phase to search around the newly generated solutions. 
Specifically, starting with a population of initial solutions, the algorithm repeats a number of evolution cycles (also called generations). At each generation, two parent solutions are randomly chosen from the current population. Then the recombination operator is applied to the parent solutions to generate an offspring solution which is subsequently optimized by the local search procedure. Finally, the population is updated with the improved offspring solution according to \textcolor{black}{the solution quality of the offspring.} This process is repeated until a stop condition (e.g., number of generations, computing time) is met. 

The main scheme of the memetic algorithm is described in Algorithm \ref{Main_Mem}, and the detailed descriptions of the four main components (i.e., population initialization, local search procedure, recombination operator and population updating strategy) are provided in the following subsections.

\begin{algorithm}[h]
 \begin{small}
 \caption{The Pseudo-code of the memetic algorithm}\label{Main_Mem}
 \begin{algorithmic}[1]
 \STATE \textbf{Input}: The graph $G$
 \STATE \textbf{Output}: The best solution found so far
 \STATE $P=\{x^1, x^2, \dots, x^p\} \leftarrow$ randomly generate $p$ initial solutions    /$*$ Section \ref{Initial population} $*$/
 \STATE \textbf{for} $i=1,2,\dots,p$ \textbf{do}
 \STATE     \ \ \  $x^i \leftarrow$ Local\_Search($x^i$)    \ \ \    /$*$ Section \ref{subsec_local search} $*$/
 \STATE \textbf{end for}
 \REPEAT
    \STATE     Randomly choose two individuals $x^a$ and $x^b$ from $P$
    \STATE      $x^0$ $\leftarrow$ Recombination($x^a$, $x^b$) \ \ \   /$*$ Section \ref{combine_operator} $*$/
    \STATE      $x^0\leftarrow$ Local\_Search($x^0$)    \ \ \  /$*$ Section \ref{subsec_local search} $*$/
    \STATE  $ P \leftarrow$ Pool\_Updating ($x^0, P$)  \ \ \ /$*$ Section \ref{pool_update} $*$/
 \UNTIL{stop condition is met}
 \RETURN{the best solution found so far}
 \end{algorithmic}
 \end{small}
\end{algorithm}

\subsection{Initial Population}
\label{Initial population}

The initial population is generated as follows. A random permutation is first created and then improved by the local search procedure (see Section \ref{subsec_local search}). If the improved solution is not already present in the population, it is added into the population. Otherwise, this solution is discarded and a new random permutation is created. This procedure is iterated until the population is filled with $p$ solutions ($|P|=p$ is the population size).

\subsection{Local Search Procedure}
\label{subsec_local search}

Given a solution $x$ (i.e., a permutation), we generate a neighboring solution by applying to $x$ an operator called \textsl{insert}, which moves a vertex from its current position $i$ to another position $j$, denoted by \textsl{insert}$(i,j)$. This operator is widely used in the \textsl{classical} linear ordering problem (LOP) (see e.g. \cite{LOP_Scatter,LOP_VNS}).

\textcolor{black}{Our local search procedure is inspired by \textit{forward} local search in \cite{EvPR}, and is divided into two parts: \textsl{forward} and \textsl{backward}. Instead of employing the traditional method (select the best solution from the whole neighborhood, which is very time-consuming), the local search procedure considers the vertex (denoted by $\textsl{v}^{*}$) with the maximum $\alpha$ value. We assume that the position of $\textsl{v}^{*}$ is $pos_{v^{*}}$. The set of target positions ($T_{pos}$) is composed of positions before $pos_{v^{*}}$. By employing \textsl{insert} move, $\textsl{v}^{*}$ is moved from the current position to best target position chosen from $T_{pos}$ with respect to the objective function value. If there is no improving move associated with $\textsl{v}^{*}$, we turn to the next element with the maximum $\alpha$ value. This process is called the \textsl{forward} part of our local search procedure, which is used in \cite{EvPR}.}

\textcolor{black}{It should be noted that this procedure is much faster than the traditional local search method. The \textsl{forward} part is repeated until the current solution cannot be further improved. The \textsl{backward} part, which is opposite with the \textsl{forward} part, selects the vertex with the minimum $\alpha$ value. The set of target positions ($T_{pos}$) includes positions after the position of the selected vertex. The \textsl{backward} part is also repeated until the current solution cannot be further improved. This \textsl{backward} part stops when it cannot further improve the solution quality any more. Note that this \textsl{backward} part is missed in previous papers but it has significant influence on both solution quality and computational efficiency according to our experiments (see Section \ref{Sec_Analysis}). The details of  local search procedure is described in Algorithm \ref{Local_Search}.}

To accelerate the evaluation of neighboring solutions, we employ a fast incremental evaluation technique introduced for LOP in \cite{LOP}. The main idea is to maintain a special data structure to record the move values for swapping adjacent vertices. Particularly, each \textsl{insert} move is decomposed into several \textsl{swap} moves, which sequentially exchanges the vertices in adjacent positions. Experiments show that this method can reduce the computational time of the local search procedure by about 65\%.

\renewcommand{\baselinestretch}{1.0}\large\normalsize
 \begin{algorithm}[h]
 \begin{small}
 \caption{The pseudo-code of the local search procedure}\label{Local_Search}
 \begin{algorithmic}[1]
 \STATE \textbf{Input}:  An initial solution $x^0$
 \STATE \textbf{Output}: A locally optimal solution
 \STATE improved $\leftarrow$ true
 \REPEAT
    \STATE improved $\leftarrow$ false
    \FOR{$i=1, 2, 3, \dots, n$ }
        \STATE Figure out the vertex $v^{*}$ with the $i$th large $\alpha$
        \STATE Let $pos_{v^{*}}$ be the position of $v^{*}$
            \FOR{$j=pos_{v^{*}}-1, \dots, 1$}
                \STATE Swap the vertex on $j$ and $j+1$
                \STATE Calculate the objective function value $f^{'}$
                \IF{$f^{'} \leqslant f_{best}$}
                    \STATE $pos_{best}$ $\leftarrow$ $j$ and improved $\leftarrow$ true
                    \STATE $f_{best}$ $\leftarrow$ $f^{'}$
                \ENDIF
            \ENDFOR
        \STATE Move vertex $v^{*}$ to position $pos_{best}$
    \ENDFOR
 \UNTIL{improved = false}
 \STATE improved $\leftarrow$ true
 \REPEAT
    \STATE improved $\leftarrow$ false
    \FOR{$i=1, 2, 3, \dots, n$ }
        \STATE Figure out the vertex $v^{*}$ with the $i$th small $\alpha$
        \STATE Let $pos_{v^{*}}$ be the position of $v^{*}$
            \FOR{$j=pos_{v^{*}}+1, \dots, n$}
                \STATE Swap the vertex on $j$ and $j-1$
                \STATE Calculate the objective function value $f^{'}$
                \IF{$f^{'} \leqslant f_{best}$}
                    \STATE $pos_{best}$ $\leftarrow$ $j$ and improved $\leftarrow$ true
                    \STATE $f_{best}$ $\leftarrow$ $f^{'}$
                \ENDIF
            \ENDFOR
        \STATE Move vertex $v^{*}$ to position $pos_{best}$
    \ENDFOR
 \UNTIL{improved = false}
 \end{algorithmic}
 \end{small}
\end{algorithm}
\renewcommand{\baselinestretch}{1.5}\large\normalsize
\subsection{Recombination Operator}
\label{combine_operator}
Previous studies suggest that the recombination operator is a key ingredient that determines the performance of the memetic algorithm. In this paper, we adopt the order-based operator which has been proven to be very useful for the classical LOP problem \cite{LOP}.

Specifically, two solutions in the current population are randomly selected as parent solutions. To maintain the diversity of the population, we restrict that the \textsl{similarity} (see Section \ref{Sec_Analysis}) between the parent solutions should not be smaller than the average \textsl{similarity} of the solutions in the population.

Then, the order-based recombination operates in two phases. First, we copy one of the parents to the offspring solution. Second, we randomly select $k$ positions of the offspring solution and reorder the vertices on these $k$ positions according to their orders in another parent. Here, we set experimentally $k=n/2$ ($n$ being the number of vertices of the problem instance).

For example, we assume that $x^a=(2,3,1,4,6,5)$ and $x^b=(4,1,2,5,6,3)$ are the two parent solutions and 2,4,6 are respectively the selected positions. Then, the offspring solution obtained by the order-based recombination operator is $x^0=(2,4,1,5,6,3)$.

\subsection{Population Updating}
\label{pool_update}

\textcolor{black}{The population updating strategy we employed is widely used in the literature (denoted by \textsl{PoolWorst}). This updating technique replaces the worst solution in the population with the new offspring solution if the offspring solution is good enough in terms of the objective value. Compared with some other updating strategies which consider both the solution quality and the diversity of the population, e.g., \cite{UBQP,Lu2010b,Porumbeletal2010}, \textsl{PoolWorst} strategy is able to accelerate the convergence speed, enabling the algorithm to find good solutions in relatively short time. Since our local search procedure is able to reach a balance between the search efficiency and effectiveness, it is more suitable for the updating strategy to accelerate the convergence speed.}

\section{Computational Results and Comparison}
\label{Sec_Results}
In this section, we assess the performance of the memetic algorithm on two sets of benchmark instances and compare it with the state-of-the-art algorithms in the literature.

\subsection{Problem Instances and Experimental Protocol}
\label{subsec_reference_alg}
Two sets of instances are used in the experiments. The first set of instances called LOLIB consists of $43$ small instances with $n=44$ to $60$. This set of instances is well known for the LOPCC problem. The second set of instances named RANDOM includes $75$ instances. Specifically, $25$ of them are small instances of size $n=35$, and the other two parts are large and difficult instances of size $n=100$ and $n=150$. These instances are widely-used in a number of studies, see for example \cite{TS,EvPR}.

The proposed algorithm is programmed in C++ and compiled using Microsoft Visual Studio 6.0 on a PC running Windows XP with \textcolor{black}{2.4 GHz CPU and 2 Gb RAM}. \textcolor{black}{  The population size is set to $15$ for all the tested instances. The stop condition for the algorithm is a fixed number of generations. For the instances of size $n=150$, the generation limit is set to $200$. For all the other instances, the generation limit is set to $100$. Given the stochastic nature of the memetic algorithm, each instance is solved independently 10 times using different random seeds (some large instances are solved $5$ times). } 


\subsection{Computational Results}
\label{subsec_results}
  \textcolor{black}{ Tables \ref{tab:lolib} to \ref{tab:rand150b} summarize the computational statistics of the memetic algorithm. In each table, column 1 gives the instance name. Columns $2$ and $3$ respectively give the number of vertices ($n$) and the previous best-known objective values ($f_{prev}$). Columns $4-6$ report the statistics of the memetic algorithm: the best found objective value ($f_{best}$), the gap between our best results and the previous best-known results ($g_{best}$), the number of generations needed to reach the given $f_{best}$ (\emph{iter}). If there are multiple runs reaching the best objective value, \emph{iter} stands for the average number of needed generations. Particularly, the number ``0'' means that the best found result is obtained during the initialization process. Column (\emph{$time_{best}$}) gives the average computational time in seconds to detect the best objective values. The last column (\emph{$time_{total}$}) shows the average total execution time in seconds. The last row named $Average$ gives the average value for each column.}

\renewcommand{\baselinestretch}{1.2}\large\normalsize
\begin{center}
\begin{table}
\begin{scriptsize}
\caption{\textcolor{black}{Computational results on the 43 LOLIB instances}}
\label{tab:lolib}
\centering{
\begin{tabular}{p{1.5cm}|p{0.8cm}p{2.5cm}p{2.5cm}p{1.5cm}p{0.7cm}p{0.9cm}p{0.9cm}p{0.01cm}}\hline& &  & \multicolumn{5}{l}{Memetic Algorithm} &\\
\cline{4-8}
{\raisebox{2.50ex}[0cm][0cm]{Instance}} &  {\raisebox{2.50ex}[0cm][0cm]{$n$}} &  {\raisebox{2.50ex}[0cm][0cm]{$f_{prev}$}} & {$f_{best}$} & {$g_{best}$} & {\textsl{iter}} & {\textsl{$time_{best}$}} &{\textsl{$time_{total}$}}&
\\
\hline
{be75eec}    & {50} & {5.085} & {5.085} & {0}  & {100} & {8.53}    & {12.83}    &\\
{be75np}    & {50} & {16543433.910} & {16543433.910}& {0}  & {91} & {9.13}& {12.90}    &\\
{be75oi}    & {50} & {2.788}& {2.788} & {0}  & {146} & {10.16}&  {11.77}    &\\
{be75tot}    & {50} & {297138.237} & {297139.434}& {1.197}  & {29} & {3.84}  &{8.46}    &\\
{stabu1}    & {60} & {13.284}& {13.284}& {0}  & {137} & {23.84} & {27.35}    &\\
{stabu2}    & {60} & {14.029} & {14.029}& {0}  & {89} & {20.03}  & {29.76}    &\\
{stabu3}    & {60} & {9.412} & {9.412}& {0}  & {167} & {26.35}&  {32.38}    &\\
{t59b11xx}    & {44} & {76261.813} & {76261.815}& {0.002}  & {38} & {3.02}& {6.93}    & \\
{t59d11xx}    & {44} & {4086.303} & {4086.303}& {0}  & {125} & {4.76} &{6.79}    &\\
{t59n11xx}    & {44} & {1618.897} & {1618.907}& {0.010}  & {90} & {5.12} & {8.23}    &\\
{t65b11xx}    & {44} & {28230.444} & {28230.444}& {0}  & {37} & {4.04} & {8.35}    &\\
{t65d11xx}    & {44} & {3898.568} & {3898.568}& {0}  & {53} & {5.26} & {9.06}    & \\
{t65i11xx}    & {44} & {826108.100} & {826108.100}& {0}  & {55} & {3.91} & {8.75}    &\\
{t65l11xx}    & {44} & {2657.735}& {2657.735}& {0}  & {13} & {2.34} & {9.75}    &\\
{t65w11xx}    & {44} & {19.258} & {19.258}& {0}  & {52} & {4.76}  & {8.56}    &\\
{t69r11xx}    & {44} & {14.036}& {14.037}& {0.001}  & {8} & {1.83} & {7.66}    &\\
{t70b11xx}    & {44} & {93.671}& {93.671}& {0}  & {0} & {0.76} & {8.50}    &\\
{t70d11xn}    & {6} & {1.80}& {8.93}& {0}  & {6} & {1.82} &{8.93}    & \\
{t70d11xx}    & {44} & {4.435}& {4.435}& {0}  & {17} & {2.04} & {6.75}    &\\
{t70f11xx}    & {44} & {1.267}& {1.267}& {0}  & {69} & {3.03} & {5.94}    &\\
{t70i11xx}    & {44} & {121146.029}& {121146.576}& {0.547}  & {8} & {1.51} & {7.07}    &\\
{t70k11xx}    & {44} & {0.492}& {0.492}& {0}  & {55} & {3.56} & {7.32}    &\\
{t70l11xx}    & {44} & {798.919}& {798.923}& {0.004}  & {6} & {1.17} & {6.02}    &\\
{t70u11xx}    & {44} & {35490330260.185}& {35490477444.038}& {147183.852}  & {16}    & {1.63} & {6.63} &\\
{t70x11xx}    & {44} & {0.231}& {0.231}& {0}  & {5} & {1.07} &  {5.03}    &\\
{t74d11xx}    & {44} & {4.756} & {4.757}& {0.001}  & {53} & {3.63} & {6.41}    &\\
{t75d11xx}    & {44} & {5.059} & {5.059}& {0}  & {19} & {2.01} & {5.16}    &\\
{t75e11xx}    & {44} & {2062.289} & {2062.289}& {0}  & {19} & {2.10} & {6.76}    &\\
{t75i11xx}    & {44} & {4454.931} & {4454.931}& {0}  & {0} & {0.72} & {7.41}    &\\
{t75k11xx}    & {44} & {1.323} & {1.323}& {0}  & {63} & {3.36} & {6.62}    &\\
{t75n11xx}    & {44} & {9.897}& {9.897}& {0}  & {61} & {3.12} & {5.61}    &\\
{t75u11xx}    & {44} & {0.326} & {0.326}& {0}  & {0} & {0.55}  &{5.61}    &\\
{tiw56n54}    & {56} & {2.645}& {2.645}& {0}  & {1} & {3.42} &{19.85}    &\\
{tiw56n58}    & {56} & {3.620}& {3.620}& {0}  & {85} & {12.43} & {17.1}    &\\
{tiw56n62}    & {56} & {3.024}& {3.024}& {0}  & {79} & {10.63} & {20.07}    &\\
{tiw56n66}    & {56} & {2.687}& {2.687}& {0}  & {5} & {3.27} &{16.87}    &\\
{tiw56n67}    & {56} & {1.877}& {1.877}& {0}  & {45} & {7.54} & {15.43}    &\\
{tiw56n72}    & {56} & {1.567}& {1.567}& {0}  & {47} & {13.12} & {21.51}    &\\
{tiw56r54}    & {56} & {2.626}& {2.626}& {0}  & {80} & {18.35} & {24.42}    &\\
{tiw56r58}    & {56} & {3.602}& {3.602}& {0}  & {53} & {12.80} & {22.54}    &\\
{tiw56r66}    & {56} & {2.189}& {2.189}& {0}  & {136} & {17.48} &{22.61}    &\\
{tiw56r67}    & {56} & {1.541}& {1.541}& {0}  & {64} & {14.87} &{26.61}    &\\
{tiw56r72}    & {56} & {1.349}& {1.349}& {0}  & {86} & {14.97} &{23.17}    &\\
\hline
{Average}   & {48.7} & {825773080.516} & {825776503.437} & {79.603}   & {36} & {6.92} &{12.58}    &\\
\hline
\end{tabular}}
\end{scriptsize}
\end{table}
\end{center}
\renewcommand{\baselinestretch}{1.5}\large\normalsize

\renewcommand{\baselinestretch}{1.5}\small\normalsize
\begin{table}
\begin{scriptsize}
\caption{\textcolor{black}{Computational results  on the 25 RANDOM instances with $n=35$}}
\label{tab:rand35}

\begin{tabular}{p{1.5cm}|p{1.1cm}p{1.9cm}p{1.9cm}p{1.4cm}p{1.1cm}p{1.1cm}p{0.9cm}p{0.01cm}}
\hline
&  &  & \multicolumn{5}{l}{Memetic Algorithm} & \\
\cline{4-8}
{\raisebox{2.50ex}[0cm][0cm]{Instance}} &  {\raisebox{2.50ex}[0cm][0cm]{$n$}} &  {\raisebox{2.50ex}[0cm][0cm]{$f_{prev}$}} & {$f_{best}$} & {$g_{best}$} & {\textsl{iter}} & {\textsl{$time_{best}$}}& {\textsl{$time_{total}$}}&
\\
\hline
{t1d35.1}    & {35} & {0.923} & {0.923} & {0}  & {44} & {1.33}    &{3.14}    &\\
{t1d35.2}    & {35} & {0.167} & {0.167}& {0}  & {27} & {1.76}& {1.96}    &\\
{t1d35.3}    & {35} & {0.154}& {0.154} & {0}  & {7} & {0.54}& {2.01}    & \\
{t1d35.4}    & {35} & {0.196} & {0.196}& {0}  & {41} & {1.06}  &{2.03}    &\\
{t1d35.5}    & {35} & {1.394}& {1.394}& {0}  & {14} & {0.69} & {1.87}    &\\
{t1d35.6}  &{35} & {0.200} & {0.200}& {0}  &  {67}  &{1.27} & {2.33}  & \\
{t1d35.7}    & {35} & {0.120} & {0.120}& {0}  & {29} & {0.96}& {2.74}    & \\
{t1d35.8}    & {35} & {0.226} & {0.226}& {0}  & {11} & {0.79}&  {2.03}    &\\
{t1d35.9}    & {35} & {0.436} & {0.436}& {0}  & {62} & {1.27} &{2.23}    &\\
{t1d35.10}    & {35} & {0.205}& {0.223}& {0.018}  & {6} & {0.51} & {1.67}    & \\
{t1d35.11}    & {35} & {0.369} & {0.369}& {0}  & {0} & {0.43} &{0.58}    & \\
{t1d35.12}    & {35} & {0.234} & {0.234}& {0}  & {38} & {1.03} & {2.07}    &\\
{t1d35.13}    & {35} & {0.196} & {0.196}& {0}  & {9} & {0.56} &  {1.87}    &\\
{t1d35.14}    & {35} & {0.138} & {0.138}& {0}  & {37} & {0.94} & {1.93}    &\\
{t1d35.15}    & {35} & {1.376}& {1.376}& {0}  & {9} & {0.64} &{2.21}    & \\
{t1d35.16}    & {35} & {0.286} & {0.286}& {0}  & {11} & {0.67}  &{2.12}    & \\
{t1d35.17}    & {35} & {0.199}& {0.199}& {0}  & {1} & {0.48} & {2.99}    &\\
{t1d35.18}    & {35} & {0.381}& {0.381}& {0}  & {7} & {0.57} &{2.43}    & \\
{t1d35.19}    & {35} & {0.236}& {0.236}& {0}  & {128} & {2.14} &{2.62}    & \\
{t1d35.20}    & {35} & {0.068}& {0.068}& {0}  & {43} & {1.13} &{2.23}    & \\
{t1d35.21}    & {35} & {0.202}& {0.202}& {0}  & {10} & {0.87} & {2.76}    &\\
{t1d35.22}    & {35} & {0.177}& {0.177}& {0}  & {5} & {0.43} & {1.77}    &\\
{t1d35.23}    & {35} & {0.345}& {0.345}& {0}  & {28} & {0.94} &{2.06}    & \\
{t1d35.24}    & {35} & {0.132}& {0.132}& {0}  & {12} & {0.53} & {1.7}    &\\
{t1d35.25}    & {35} & {0.143}& {0.143}& {0}  & {9} & {0.67} & {1.93}    &\\
\hline
{Average}   & {35} & {0.341} & {0.341} & {0}  & {11} & {0.84} &{2.17}    & \\
\hline
\end{tabular}
\end{scriptsize}
\end{table}
\renewcommand{\baselinestretch}{1.5}\large\normalsize

\renewcommand{\baselinestretch}{1.5}\small\normalsize

\begin{table}[!ht]
\begin{scriptsize}
\caption{\textcolor{black}{Computational results  on the 25 RANDOM  instances with $n=100$}}
\label{tab:rand100}

\begin{tabular}{p{1.7cm}|p{1.0cm}p{1.3cm}p{1.5cm}p{1.5cm}p{1.0cm}p{1.0cm}p{0.9cm}p{0.01cm}}
\hline
& &  & \multicolumn{5}{l}{Memetic Algorithm} &\\
\cline{4-9}
{\raisebox{0.25ex}[0cm][0cm]{Instance}} &  {\raisebox{0.25ex}[0cm][0cm]{$n$}} &  {\raisebox{0.25ex}[0cm][0cm]{$f_{prev}$}} & {$f_{best}$} & {$g_{best}$} & {\textsl{iter}} & {$time_{best}$} & {$time_{total}$} &\\
\hline
{t1d100.1}    & {100} & {253.988} & {\textbf{246.279}} & {\textbf{-7.709}}  & {146} & {789}   & {997}  &\\
{t1d100.2}    & {100} & {288.372} & {\textbf{284.924}}& {\textbf{-3.448}}  & {196} & {990}& {1060}& \\
{t1d100.3}    & {100} & {1307.432}& {\textbf{1236.237}} & {\textbf{-71.195}}  & {142} & {771}&  {993}& \\
{t1d100.4}    & {100} & {7539.979} & {\textbf{6735.661}}& {\textbf{-804.318}}  & {196} & {913}  & {948}& \\
{t1d100.5}    & {100} & {169.336}& {\textbf{162.423}}& {\textbf{-6.913}}  & {196} & {1106} & {1136}& \\
{t1d100.6}    & {100} & {395.035} & {\textbf{391.662}}& {\textbf{-3.373}}  & {189} & {763}  & {784}& \\
{t1d100.7}    & {100} & {5936.281} & {\textbf{5641.137}}& {\textbf{-295.144}}  & {170} & {798}&  {1055}& \\
{t1d100.8}    & {100} & {2760.619} & {\textbf{2750.802}}& {\textbf{-9.817}}  & {90} & {479}&  {763}& \\
{t1d100.9}    & {100} & {62.942} & {\textbf{62.775}}& {\textbf{-0.167}}  & {177} & {805} &{917}& \\
{t1d100.10}    & {100} & {162.942}& {\textbf{159.126}}& {\textbf{-3.816}}  & {195} & {960} &  {1076}& \\
{t1d100.11}    & {100} & {233.586} & {\textbf{230.810}}& {\textbf{-2.776}}  & {112} & {567} & {996}& \\
{t1d100.12}    & {100} & {236.696} & {\textbf{231.176}}& {\textbf{-5.520}}  & {178} & {954} & {965}& \\
{t1d100.13}    & {100} & {593.319} & {\textbf{578.307}}& {\textbf{-15.012}}  & {187} & {888} & {919}&  \\
{t1d100.14}    & {100} & {249.162} & {\textbf{247.313}}& {\textbf{-1.849}}  & {132} & {747} &{947}&  \\
{t1d100.15}    & {100} & {406.478}& {408.312}& {1.834}  & {192} & {1081} & {1158}& \\
{t1d100.16}    & {100} & {707.413} & {707.413}& {0}  & {86} & {492}  & {983}& \\
{t1d100.17}    & {100} & {725.790}& {\textbf{718.920}}& {\textbf{-6.87}}  & {156} & {801} & {853}& \\
{t1d100.18}    & {100} & {622.942}& {\textbf{621.940}}& {\textbf{-1.002}}  & {110} & {653} & {842}& \\
{t1d100.19}    & {100} & {228.486}& {\textbf{227.374}}& {\textbf{-1.112}}  & {130} & {674} & {1021}& \\
{t1d100.20}    & {100} & {255.151}& {\textbf{238.586}}& {\textbf{-16.565}}  & {128} & {791} & {913}& \\
{t1d100.21}    & {100} & {228.590}& {\textbf{221.462}}& {\textbf{-7.128}}  & {192} & {1131} & {1163}& \\
{t1d100.22}    & {100} & {159.336}& {\textbf{141.255}}& {\textbf{-18.081}}  & {111} & {676} & {1068}& \\
{t1d100.23}    & {100} & {1658.168}& {\textbf{1656.877}}& {\textbf{-1.291}}  & {196} & {881} & {935}& \\
{t1d100.24}    & {100} & {469.658}& {\textbf{468.863}}& {\textbf{-0.795}}  & {180} & {875} &{1091}&  \\
{t1d100.25}    & {100} & {644.782}& {\textbf{637.523}}& {\textbf{-7.259}}  & {89} & {533} & {950}& \\
\hline
{Average}   & {100} & {1051.859} & {\textbf{1000.286}} & {\textbf{-51.573}}  & {155} & {268} & {327}& \\
\hline
\end{tabular}
\end{scriptsize}
\end{table}
\renewcommand{\baselinestretch}{1.5}\large\normalsize

\renewcommand{\baselinestretch}{1.5}\small\normalsize
\begin{table}
\begin{scriptsize}
\caption{\textcolor{black}{Computational results on the 25 RANDOM  instances with $n=150$}}
\label{tab:rand150b}

\begin{tabular}{p{1.5cm}|p{0.7cm}p{1.9cm}p{1.9cm}p{1.9cm}p{0.6cm}p{0.8cm}p{0.9cm}p{0.01cm}}
\hline
& &  & \multicolumn{5}{l}{Memetic Algorithm} &\\
\cline{4-9}
{\raisebox{2.50ex}[0cm][0cm]{Instance}} &  {\raisebox{2.50ex}[0cm][0cm]{$n$}} &  {\raisebox{2.50ex}[0cm][0cm]{$f_{prev}$}} & {$f_{best}$} & {$g_{best}$} & {\textsl{iter}} &{$time_{best}$} & {$time_{total}$} & \\
\hline
{t1d150.1}    & {150} & {8588.289} & {\textbf{8293.108}} & {\textbf{-295.181}}  & {102} & {1075}    &{1577}  &\\
{t1d150.2}  & {150} & {184853.686} & {\textbf{159339.130}}& {\textbf{-25514.556}} &{172} & {1598}  & {1819}  &\\
{t1d150.3}    & {150} & {574943.633}& {\textbf{548507.282}} & {\textbf{-26436.351}}  & {184} & {1604}&  {1803}  &\\
{t1d150.4}    & {150} & {75510.287} & {\textbf{68125.331}}& {\textbf{-7384.956}}  & {199} & {1779}  &{1817} & \\
{t1d150.5}    & {150} & {79069.363}& {\textbf{75426.662}}& {\textbf{-3642.701}}  & {165} & {1609} & {1920} & \\
{t1d150.6}    & {150} & {46829.985} & {\textbf{46013.112}}& {\textbf{-816.873}}  & {200} & {1688}  & {1885} & \\
{t1d150.7}    & {150} & {161149.153} & {\textbf{150146.763}}& {\textbf{-11002.390}}  & {200} & {1673}&  {1731}  &\\
{t1d150.8}    & {150} & {251940.422} & {\textbf{247564.438}}& {\textbf{-4375.984}}  & {175} & {1606}&  {1718}  &\\
{t1d150.9}    & {150} & {364320.250} & {\textbf{363221.346}}& {\textbf{-1098.904}}  & {197} & {1627} &{1675}  &\\
{t1d150.10}    & {150} & {122217.421}& {\textbf{107685.011}}& {\textbf{-14532.410}}  & {161} & {1500} & {1753}  & \\
{t1d150.11}    & {150} & {13900.039} & {\textbf{12360.337}}& {\textbf{-1539.702}}  & {190} & {1544} & {1729}  &\\
{t1d150.12}    & {150} & {65717.265} & {\textbf{60614.534}}& {\textbf{-5102.731}}  & {196} & {1676} & {1886} &\\
{t1d150.13}    & {150} & {109460.320} & {\textbf{105265.302}}& {\textbf{-4195.018}}  & {160} & {1403} &  {1906}  &\\
{t1d150.14}    & {150} & {74854.867} & {\textbf{70153.934}}& {\textbf{-4700.933}}  & {188} & {1786} &{1979}   &\\
{t1d150.15}    & {150} & {352880.286}& {\textbf{321468.489}}& {\textbf{-31411.797}}  & {193} & {1721} & {1779}  &\\
{t1d150.16}    & {150} & {16950196.691} & {\textbf{16915821.128}}& {\textbf{-34375.563}}  & {116} & {1114}  & {1428} & \\
{t1d150.17}    & {150} & {77828.419}& {\textbf{74903.919}}& {\textbf{-2924.500}}  & {182} & {1552} & {1786}  &\\
{t1d150.18}    & {150} & {711286.599}& {\textbf{654737.416}}& {\textbf{-56549.183}}  & {199} & {1571} & {1658} & \\
{t1d150.19}    & {150} & {67840.414}& {\textbf{66614.402}}& {\textbf{-1226.012}}  & {193} & {1590} &{1733}   &\\
{t1d150.20}    & {150} & {1886041.875}& {2074926.337}& {188884.462}  & {200} & {1907} & {1911} & \\
{t1d150.21}    & {150} & {41453.911}& {\textbf{39248.997}}& {\textbf{-2204.914}}  & {151} & {1356} & {1604}  &\\
{t1d150.22}    & {150} & {695751.688}& {\textbf{671281.287}}& {\textbf{-24470.401}}  & {139} & {1128} &{1774}   &\\
{t1d150.23}    & {150} & {22203891.826}& {\textbf{21468279.568}}& {\textbf{-735612.258}}  & {199} & {1563} &{1844}   &\\
{t1d150.24}    & {150} & {105162.367}& {\textbf{101072.915}}& {\textbf{-4089.452}}  & {194} & {1844} &{1888}   &\\
{t1d150.25}    & {150} & {462316.511}& {465798.731}& {3482.220}  & {153} & {1484} & {1703} & \\
\hline
{Average}   & {150} & {1827520.220} & {\textbf{1795074.779}} & {\textbf{-32445.444}}  & {176} & {1560} & {1772} & \\
\hline
\end{tabular}
\end{scriptsize}
\end{table}
\renewcommand{\baselinestretch}{1.5}\large\normalsize

\textcolor{black}{
Tables \ref{tab:lolib} and \ref{tab:rand35} show the computational statistics of the memetic algorithm on the small sized instances. For the 43 LOLIB instances and $25$ RANDOM instances with $n=35$, the memetic algorithm can achieve, respectively, $35$ and $24$ optimal values within a short time.}

\textcolor{black}{
Tables \ref{tab:rand100} and \ref{tab:rand150b} report the computational results  on the challenging large instances. For these instances, the memetic algorithm is able to  improve a number of the previous best known results. Specifically, for the RANDOM instances with $n=100$, the memetic algorithm improves the previous best known results for $23$  out of $25$ instances (indicated in bold). For one of the remaining instance ($t1d100.16$), the algorithm can match the previous best known result. While for the instance ($t1d100.15$), our result is worse than the previous best known one. When it comes to RANDOM instances of size $n=150$ (Table \ref{tab:rand150b}), the memetic algorithm improves the previous best known results for $23$ out of $25$ instances (indicated in bold). The results for the two remaining instances are worse than the previous best known ones.}

\textcolor{black}{
In summary, the computational results establish the efficiency and effectiveness of the proposed memetic algorithm.  For the 68 small size instances, LOLIB instances and RANDOM instances with $n=35$, the algorithm successfully matchs the best-known results on most of the instances in a short time. While for the 50 challenging large instances, RANDOM instances with $n=100$ and $n=150$, the algorithm is able to find better solutions on 46 instances within a reasonable time.}


\subsection{Comparisons with the State-of-the-art Algorithms}
\label{subsec_compare}

\renewcommand{\baselinestretch}{1.5}\large\normalsize
\begin{table}[!ht]
\begin{scriptsize}
\captionsetup{font={small}}
\caption{ \textcolor{black}{Summarized comparison of the proposed algorithm with two state-of-the-art algorithms on all the instances}}
\label{table_comparison}
\begin{tabular}{p{3.4cm}|p{2.2cm}p{1.9cm}p{1.8cm}p{1.8cm}p{0.1cm}}
\hline
{\raisebox{0.3ex}[0cm][0cm]{Instances}}& {\raisebox{0.3ex}[0cm][0cm]{}}& {\raisebox{0.3ex}[0cm][0cm]{TS}}& {\raisebox{0.3ex}[0cm][0cm]{EvPR}}& {\raisebox{0.3ex}[0cm][0cm]{Memetic}}&\\
\hline
\multirow{2}[0]{*}{43 LOLIB instances} & {Obj.function} & {8.26E+08} & {\textbf{1.35}} &  {8.26E+08} & \\
 & {Avg.deviation} & {1.40\%} & {0.00\%} &  {0.00\%} & \\
{} & {Num.of opt} & {28} & {36} & {34} & \\
{} & {Total CPU seconds} & {37.74} & {32.34} & {12.50} & \\
\hline
\multirow{2}{3.3cm}{25 random instances with $n=35$} & {Obj.function} & {0.34} & {0.34} &  {0.34} & \\
   & {Avg.deviation} & {0.51\%} & {0.45\%} &  {0.35\%} & \\
{} & {Num.of opt} & {21} & {24} & {24} & \\
{} & {Total CPU seconds} & {3.40} & {3.75} & {2.17} & \\
\hline
\multirow{2}{3.3cm}{25 random instances with $n=100$} & {Obj.function} & {1161.46} & {1058.78} &  {1034.30} & \\
& {Avg.deviation} & {16.1\%} & {5.8\%} &  {3.4\%} & \\
{} & {Num.of best} & {$\leq$2} & {$\leq$2} & {8} & \\
{} & {Total CPU seconds} & {406.79} & {351.38} & {325.45} & \\
\hline
\multirow{2}{3.3cm}{25 random instances with $n=150$} & {Obj.function} & {2.27E+06} & {1.85E+06} &  {1.90E+06} & \\
 & {Avg.deviation} & {11.64\%} & {6.03\%} &  {7.07\%} & \\
{} & {Num.of best} & {$\leq$2} & {$\leq$2} & {1} & \\
{} & {Total CPU seconds} & {2074.31} & {1127.24} & {1123.17} & \\
\hline
\end{tabular}
\end{scriptsize}
\end{table}
\renewcommand{\baselinestretch}{1.5}\large\normalsize

\textcolor{black}{In this section, the proposed memetic algorithm is compared with two state-of-the-art algorithms in the literature, respectively named TS \cite{TS} and EvPR \cite{EvPR}. The experiments are done on all the $118$ instances. We try our best to conduct the comparison on a time-equalized basis,  so we set the generation limit (the stop condition) to be $110$ for RANDOM instances with $n=150$, and $200$ for the rest instances, and run the proposed algorithm on each instance only once. Nevertheless, given the difference of computing platforms used for the compared algorithms, it is very difficult to replicate the exact stop conditions as those used in the reference algorithms. As a consequence, the comparisons shown in this section are for indicative purposes only and should be interpreted with caution.}

\textcolor{black}{Table \ref{table_comparison} presents the results of these comparisons. Column 1 gives the instance name. Column 2 presents the four comparative criteria: the average objective value (Obj.function), average deviation w.r.t. the best-known result (Ave.deviation), the number of instances where the algorithm can reach the optimal or best-known solution (Num.of best), and the average elapsed time on each instances (Total CPU seconds) . Columns 3-5 respectively present the results obtained by TS, EvPR and memetic algorithm.}

\textcolor{black}{As one can observe from Table \ref{table_comparison}, when comparing with  TS algorithm,  memetic algorithm yields better results in general. One each set of instances, the Obj.function, Ave.deviation and Num.of best of the memetic algorithm are better than the TS algorithm. The Total CPU seconds of the two algorithms on each set of instances are at the same level.}

\textcolor{black}{When comparing the memetic algorithm with the EvPR algorithm, one can observe that the performance of the two algorithms are roughly the same on the first two sets of small instances. For the 25 RANDOM instances with $n=100$, memetic algorithm's performance is better. Its Obj.function and Avg.deviation are smaller than EvPR. However, when it comes to the 25 RANDOM instances with $n=150$, memetic algorithm's performance is worse than EvPR. The reason maybe that, in order to conduct the comparison on a time-equalized basis, we set the generation limit of the memetic algorithm on these instances to be 110, and within this time period, the memetic algorithm still focuses on exploration and therefore the computational results are relatively poor. These results show that the memetic algorithm may has some complementary feature w.r.t. the EvPR algorithm. If in some applications, the allowed computation time is limit, one should choose EvPR algorithm. Otherwise, if one focuses on finding high-quality solution and allows to use relatively longer computation time, choosing memetic algorithm may be better.}

\section{ Analysis of the Forward-Backward Local Search Strategy}
\label{Sec_Analysis}

\textcolor{black}{Local search is a critical component of the memetic algorithm.  Our local search procedure is inspired by the local search procedure proposed in \cite{EvPR}. However, the local search procedure in \cite{EvPR} allows only forward moves, our local search procedure allows both $forward$ and $backward$ moves. In order to investigate whether this difference makes a meaningful contribution, we conduct experiments on some representative instances.}

\textcolor{black}{We keep other ingredients unchanged, and run two memetic algorithms, one with $forward$ local search and the other with $forward$-$backward$ local search, on  $5$ random instances (t1d100.1 t1d100.2 t1d100.11 t1d100.12 t1d100.21 t1d100.22). To reduce the impact of randomness, each algorithm is run 20 independent times. The stopping criterion is the execution time which is fixed to be $400s$.  We observe two characteristics : the average solution quality in the population and the population diversity.}

\textcolor{black}{Given two solutions $x^i$ and $x^j$, we define the \textit{distance} between $x^i$ and $x^j$ (denoted by $d_{i,j}$) as: $n$ (the number of vertexes) $-$ the length of the longest common subsequence of $x^i$ and $x^j$. Then, the population diversity (denoted by $pdi$) is defined as:}
\begin{equation}
pdi=\frac{\sum_{i=1}^{p-1} \sum_{j=i+1}^{p} (d_{i,j})}{p(p-1)/2}
\end{equation}
\textcolor{black}{With this definition, $pdi=0$ implies that all the solutions are the same. With the value of $pdi$ becoming smaller, the population diversity decreases, implying that solutions in the population tend to become more similar.}

\begin{figure}[htbp]
   \includegraphics[width=1\textwidth, angle=0]{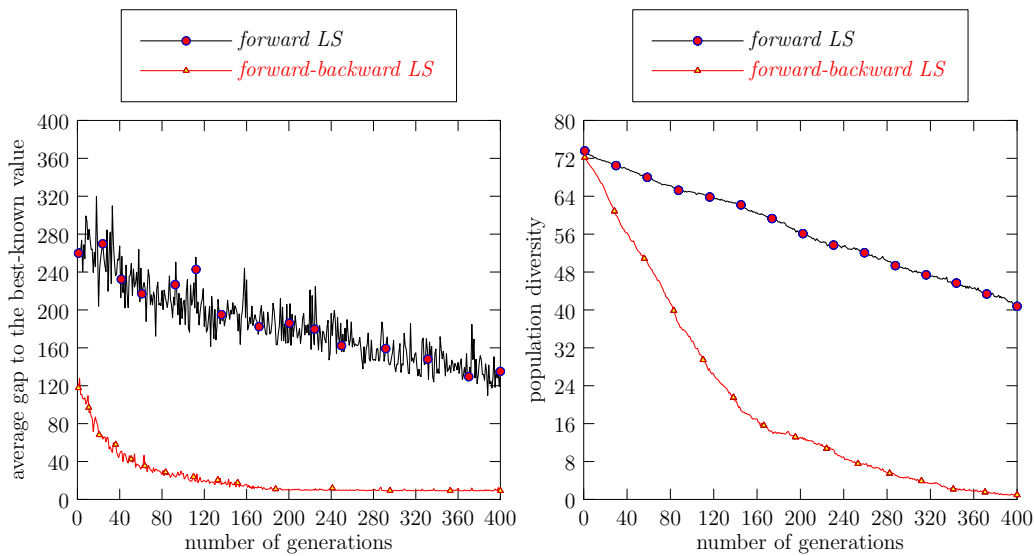}
   \caption{\textcolor{black}{Comparison between two different local search strategies}} \label{fig_neighbourhood1}
\end{figure}

\textcolor{black}{Fig.\ref{fig_neighbourhood1} shows how the average solution quality (left) and the diversity of the population (right) evolving with the execution time. As Fig. \ref{fig_neighbourhood1} shows, the algorithm with forward-backward LS can usually find better solutions, and meanwhile its convergence speed is faster.}

\section{Conclusion}
\label{Sec_Conclusions}
In this paper, we have presented a memetic algorithm for the linear ordering problem with cumulative costs. The proposed algorithm uses an order-based recombination operator for generating new solutions and an effective local search procedure for local optimization. The proposed algorithm was evaluated on 118 well-known benchmark instances and the results proved its effectiveness and efficiency. For the 43 LOLIB instances and 25 small random instances, the algorithm reaches 35 and 24 optimal values within a short computational time. In addition, it is able to improve the previous best known objective values for $46$ out of $50$ large instances.

\section*{Acknowledgments}
We are grateful for comments by the referees for their comments that help us improve the paper. This work was partially supported by the National Natural Science Foundation of China (Grant No. 61100144) and the RaDaPop (2009-2013) and LigeRo projects (2009-2013) from Pays de la Loire Region, France.


\end{document}